\documentclass[sn-mathphys,Numbered]{sn-jnl}

\usepackage{graphicx}%
\usepackage{multirow}%
\usepackage{amsmath,amssymb,amsfonts}%
\usepackage{amsthm}%
\usepackage{mathrsfs}%
\usepackage[title]{appendix}%
\usepackage{xcolor}%
\usepackage{textcomp}%
\usepackage{manyfoot}%
\usepackage{booktabs}%
\usepackage{algorithm}%
\usepackage{algorithmicx}%
\usepackage{algpseudocode}%
\usepackage{listings}%
\usepackage{subcaption}
\usepackage{lineno}
\usepackage{color,soul}

\begin{document}

\title[Article Title]{Depth Over RGB: Automatic Evaluation of Open Surgery Skills Using Depth Camera}

\author*[1]{\fnm{Ido} \sur{Zuckerman}}\email{ido.z@campus.technion.ac.il}
\author[2]{\fnm{Nicole} \sur{Werner}}
\author[1]{\fnm{Jonathan} \sur{Kouchly}}
\author[3]{\fnm{Emma} \sur{Huston}}
\author[3]{\fnm{Shannon} \sur{DiMarco}}
\author[2]{\fnm{Paul} \sur{DiMusto}}
\author[1]{\fnm{Shlomi} \sur{Laufer}}

\affil[1]{\orgdiv{Faculty of Data and Decision Sciences}, \orgname{Technion – Israel
Institute of Technology}, \orgaddress{\city{Haifa}, \postcode{3200003}, \country{Israel}}}

\affil[2]{\orgdiv{Department of Surgery}, \orgname{University of Wisconsin-Madison School of Medicine and Public Health}, \orgaddress{\street{600 Highland Ave}, \city{Madison}, \postcode{53792}, \state{WI}, \country{USA}}}

\affil[3]{\orgdiv{Clinical Simulation Program}, \orgname{University of Wisconsin Hospitals and Clinics}, \orgaddress{\street{600 Highland Ave}, \city{Madison}, \postcode{53792}, \state{WI}, \country{USA}}}

\abstract{\textbf{Purpose:} In this paper, we present a novel approach to the automatic evaluation of open surgery skills using depth cameras. This work is intended to show that depth cameras achieve similar results to RGB cameras, which is the common method in the automatic evaluation of open surgery skills. Moreover, depth cameras offer advantages such as robustness to lighting variations, camera positioning, simplified data compression, and enhanced privacy, making them a promising alternative to RGB cameras.

\textbf{Methods:} Experts and novice surgeons completed two simulators of open suturing. 
We focused on hand and tool detection, and action segmentation in suturing procedures. YOLOv8 was used for tool detection in RGB and depth videos. Furthermore, UVAST and MSTCN++ were used for action segmentation. Our study includes the collection and annotation of a dataset recorded with Azure Kinect.

\textbf{Results:} We demonstrated that using depth cameras in object detection and action segmentation achieves comparable results to RGB cameras. Furthermore, we analyzed 3D hand path length, revealing significant differences between experts and novice surgeons, emphasizing the potential of depth cameras in capturing surgical skills. We also investigated the influence of camera angles on measurement accuracy, highlighting the advantages of 3D cameras in providing a more accurate representation of hand movements.

\textbf{Conclusion:} Our research contributes to advancing the field of surgical skill assessment by leveraging depth cameras for more reliable and privacy evaluations. The findings suggest that depth cameras can be valuable in assessing surgical skills and provide a foundation for future research in this area.
}

\keywords{Surgical training, Computer vision, Depth Camera, Object detection, Open surgery}

\maketitle

\section{Introduction}\label{Introduction}
The complexity and high-stakes nature of open surgery necessitates the development of reliable and robust systems for evaluating surgical skills \cite{reznick2006teaching}. The evaluation of surgical skills has been an active area of research, with methodologies ranging from subjective assessments by expert surgeons to objective metrics using sensors and data analytics. 

Studies have shown the capability of motion sensors to distinguish between expert and novice surgeons. For instance, novices tend to move their hands with less efficiency, resulting in longer path lengths \cite{dosis2005synchronized}. Additionally, they exhibit slower movements \cite{smith2002motion} and employ a more expansive working volume \cite{d2016working}. Unfortunately, hand sensors come with drawbacks such as high costs and discomfort. Furthermore, their integration into the operating room environment poses significant challenges.

The combination of RGB cameras and computer vision provides a new approach for assessing surgical skills. Goldbraikh et al. \cite{goldbraikh2022video} utilized a standard webcam in combination with object detection to track hand movements, showing significant differences between students and experts.
This technique paves the way for the creation of simple and accessible training systems, providing learners with the opportunity to practice independently and receive objective feedback. However, motion captured by an RGB camera is the 2D in-plane of the camera rather than the actual 3D distance. The measurements can vary significantly if the camera's angle in relationship to the suture area changes. This study aims to investigate the potential of depth cameras to address this limitation, proposing a method that not only resolves this issue but also preserves the simplicity and accessibility of the training systems.

The use of RGB cameras is not limited to object detection and motion analysis, In recent years deep learning techniques have been used for general tasks such as tool detection in laparoscopic surgeries \cite{fathabadi2021multi} or surgical gesture recognition \cite{goldbraikh2022bounded}. Additionally, other studies have harnessed computer vision to formulate task-specific performance metrics \cite{halperin2023automatic, bkheet2023using}. Therefore, RGB cameras may have a broad impact on the quality, efficiency, and safety of surgical procedures. 

Nevertheless, using RGB cameras, especially in a clinical scenario, poses several challenges. First, privacy concerns emerge due to factors like capturing facial details and text. Second, lighting in the operating room is very challenging \cite{DASCALAKI2009551}, as there is a wide variation in the amount of light in different areas \cite{likitlersuang2017views,haque2020illuminating}. 
Depth cameras have been suggested as an alternative to RGB cameras to overcome these issues \cite{9795869,yeung2019computer}. They require no contact with the operating environment while still being capable of accurately tracking real-life hand motion data. They may be used to measure pose estimation and gait analysis \cite{martinez2021ethical} as well as patient activity recognition \cite{siddiqi2021unified}.

This study introduces an approach that employs depth cameras to automatically evaluate open surgery skills, specifically focusing on hand and tool detection and action segmentation in suturing procedures. 
We show that depth cameras can achieve comparable results to RGB cameras in a more robust way and provide an alternative approach for assessing surgical skills. 
The paper's main contributions are : (1) Demonstrating that depth cameras are as effective as RGB cameras for object detection and action segmentation. (2) Analysing how the angle between the camera and suture area can affect the accuracy of their results, thus demonstrating the advantage of depth data. (3) Introducing a novel metric that relies solely on depth cameras. 

\section{Methods}\label{methods}
\subsection{The Dataset}
The study included 28 participants:  22 first-year surgical residents (8 male and 14 female) and 6 attending surgeons (3 male and 3 female) at a Midwestern academic hospital.
The residents participated in this study as part of an annual surgical intern simulation series in which all first-year surgical residents complete a selection of basic surgical skills. Each participant was informed of the research prior to the session, and their decision to participate had no influence on the simulation series.
One week before the simulation series, each intern received a video showing a faculty member accurately performing each skill. There was no limit on the amount of video views. During the simulation series, each intern is paired with a faculty member in a room within the hospital simulation center. The intern is then given standardized written instructions with scoring metrics and asked to complete each skill using a simulator. After task completion, the faculty member provides feedback to the resident.

The participants were engaged in conducting various surgical tasks utilizing two simulators: a ``Suture pad'' and a ``Fascia Closure''. The execution of these tasks was documented through an Azure Kinect, which features a 4K RGB camera, a Depth Camera, and an IR Camera.

\begin{figure}[ht]
  \centering
  \begin{minipage}[b]{0.4\textwidth}
    \includegraphics[width=\textwidth]{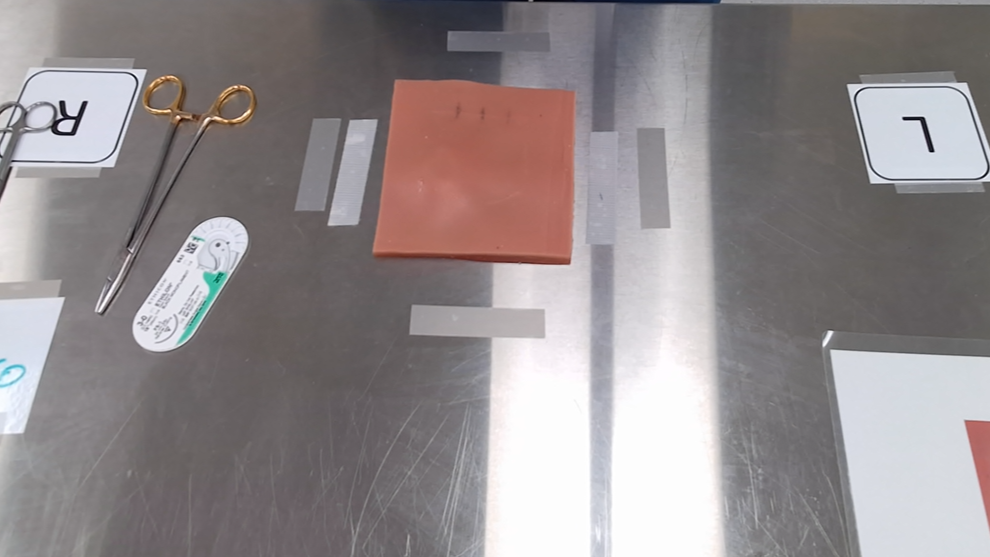}
    \caption{Suture Pad}
    \label{fig:Simulator1}
  \end{minipage}
  \hfill
  \begin{minipage}[b]{0.4\textwidth}
    \includegraphics[width=\textwidth]{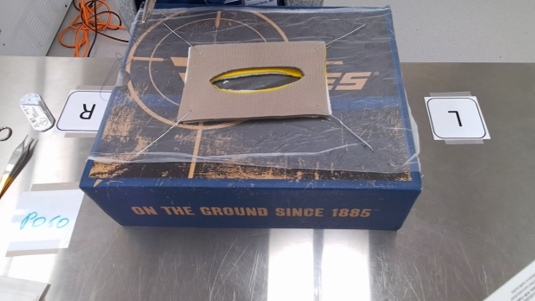}
    \caption{Fascia Closure}
    \label{fig:Simulator2}
  \end{minipage}
\end{figure}

The first simulator, the ``Suture Pad'' simulator \ref{fig:Simulator1}, was made of silicone. It was constructed to resemble human tissue and allows trainees to practice basic suturing techniques, such as creating knots and closing incisions. This simulator is similar to the simulator presented in \cite{williams2020development}. In this study, participants executed four tasks using this simulator: simple suture, horizontal mattress suture, vertical mattress suture, and running suture. The goal was to train and assess medical professionals in the technique of suturing wounds. The initial task averaged 54 seconds, the second task 84 seconds, the third 81 seconds, and the final task approximately 206 seconds.

The second simulator, the ``Fascia Closure'' simulator \ref{fig:Simulator2}, simulated the process of suturing and closing the connective tissue layer called fascia during surgical procedures. 
This simulator design is taken from Mayo Clinic’s Surgical Olympics where it has been used since 2006 \cite{buckarma2016lowcost}.
It provides a simulation of the fascia layer, enabling trainees to practice the skills required for successful closure. This simulator focuses on running suture in a distinct type of tissue. This simulator took an average of 379 seconds.

We focused on several computer vision tasks: object detection of the hands and tools as well as temporal segmentation of the surgical gestures. For object detection, about 900 frames from simulator 1 and an additional 900 frames from simulator 2 were annotated. 
These frames were drawn from 14 participants, who were chosen randomly, at a rate of one frame every 5 seconds. The number of participants was limited to 14 due to constraints on labeling resources. 
In each frame, all present tools, hands, and the simulator itself were marked with bounding boxes. The tools identified include Needle Driver, Tissue Forceps, Dressing Forceps, and Scissors. Additionally, the entire video set was annotated for temporal segmentation. The catalog of gestures consists of ``Holding needle with a tool'', ``Needle passing'', ``Pull the suture'', ``Instrumental tie'', ``Lay the knot'', ``Cut the suture'', ``No Gesture'', and ``Hand tie''.

We converted the depth matrix from each frame of the depth camera videos into grayscale video to better visualize and analyze the spatial information. In this format, objects nearest to the camera are represented in white, while those at a greater distance appear black, thereby simplifying the representation of depth information. The annotations applied to the RGB videos were also used for these depth videos. 

\subsection{Hardware and Software}

We conducted all our experiments, including training, testing, and evaluations, using a hardware setup consisting of two NVIDIA RTX A6000 GPUs and a single Intel Core i9-10940X CPU equipped with 28 logical cores. 
For running these experiments, we employed the PyTorch library, and for experiment tracking, we utilized WANDB \cite{wandb}.

\subsection{Object Detection}
\label{methods_detection}
For the purpose of object detection, encompassing tools, hands, and the simulator, the YOLOv8 architecture \cite{Jocher_YOLO_by_Ultralytics_2023} was employed. This architecture was trained using the Ultralytics framework. 
Notably, four distinct models were trained for each simulator scenario: one for RGB data encompassing all tools, another for depth data encompassing all tools, a third for RGB data focusing solely on hands, and a fourth for depth data dedicated to hands. 
In the hand-focused models, we had two classes: ``Right Hand'' and ``Left Hand'', While, for the models aimed at detecting all tools, we used: ``Right Hand'', ``Left Hand'', ``Needle Driver'', ``Tissue Forceps'', ``Dressing Forceps'', ``Scissors'', and ``Simulator''.

During the training process, several data augmentation techniques were applied to enhance the model's robustness, including rotations, and brightness adjustments.

Modifications were made to the prediction head of the model to accommodate the aforementioned classes. The evaluation of the model's performance was carried out using the mean average precision (mAP) based on intersection over union (IoU). Finally, the trained model was applied to extract per-frame bounding boxes, employing a confidence threshold ranging from 0.5 to 0.95, with increments of 0.05.
Every epoch took an estimated one minute to complete, and the model was trained for a total of 300 epochs. The memory footprint was about 19GB.

\subsection{Action Segmentation}\label{method_action_segmentation}
In our action segmentation experiments, we employed two architectures: UVAST \cite{behrmann2022unified} and MSTCN++ \cite{li2020ms}.
These architectures were selected because they complement each other effectively. MS-TCN++ is a lighter and less complex model, suitable for online, real-time inference. Conversely, UVAST, being a more feature-rich and complex model, offers greater accuracy but requires longer inference times.
Both models were implemented using their original frameworks as described in the cited papers. Both architectures leveraged RGB and optical flow features, extracted using the I3D model \cite{carreira2017quo} trained on the Kinetics 400 dataset \cite{kay2017kinetics}. Specifically for depth videos, we initially converted them to greyscale and then adapted them to RGB format by triplicating each frame for compatibility with the I3D model. In terms of training, MSTCN++ was trained over 100 epochs, with each epoch averaging around 5 seconds, while UVAST was trained 600 epochs, each averaging about one minute. 
The memory footprint was about 5GB when extracting features and 3GB for model training and prediction.

The UVAST architecture incorporated the Viterbi algorithm \cite{forney1973viterbi} during the inference stage. For experiments on simulator 2, due to limited resources and the longer video durations, we limited the hypothesis space to 10,000 during inference.

In the Suture Pad simulator, the first three tasks of the suturing are closely related as they all present the execution of what we term a ``stationary knot'' — a simple suture, a horizontal mattress suture, and a vertical mattress suture. Consequently, the model was trained on a unified dataset that included these tasks, with each task separated into a distinct video. These videos were then divided into four separate train-test splits, ensuring that all videos from a single participant fell within the same split.
The results for this model will be labeled as ``Simple Suture''. For the fourth task, in the Suture Pad simulator, the results will be categorized under ``Running Suture'', while results from the Fascia Closure simulator will be designated as ``Fascia''.

We used distinct action segmentation labels for each simulator. For the suture pad simulator, labels included G0 for holding the needle with a tool, G1 for needle passing, G2 for pulling the suture, G3 for instrumental tie, G4 for laying the knot, G5 for cutting the suture, and G6 for no action. 
The fascia closure simulator employed similar labels, with the addition of G7 for hand tie.

As previously established in the literature \cite{behrmann2022unified, li2020ms}, three evaluation metrics were employed. Frame-wise accuracy, segmental edit distance, and F1@k for $k\in \{10, 25, 50\}$. Frame-wise Accuracy assesses the ratio of correctly classified frames to total frames. Segmental Edit Distance, adapted from the Levenshtein distance, compares activity segments and is normalized by the greater length between ground truth and prediction. F1@k calculates the Intersection over Union (IoU) for each segment, categorizing them as true or false positives based on a threshold \( k \).

\subsection{3D Hand Path Length}
\label{methods_3d_path}
According to \cite{chmarra2008retracting}, the 3D hand path length metric is used to evaluate surgical skills by measuring the efficiency of a surgeon's movements. Shorter, more direct paths typically indicate higher skill and experience because they reflect a surgeon's ability to perform movements more efficiently and precisely, making this metric a valuable tool for assessing and improving surgical proficiency.

To quantify the path length traversed by the hands in a three-dimensional space, we employed the object detection algorithm \ref{methods_detection} to identify the hands. Subsequently, we extracted the coordinates of the bounding box's center. 
Utilizing the depth camera provided by the Azure Kinect, we transformed the depth information into a point cloud using the Open3D (O3D) library \cite{Zhou2018}. 
The coordinates of the bounding box were then used to extract the [x,y,z] coordinates from the point cloud, representing each hand's spatial location. 
By aggregating these spatial coordinates across frames, we calculated the total path length using Euclidean distance metrics.
For the statistical analysis, we adopted the Wilcoxon rank-sum test to compare the total path lengths between the two groups (experts and residents). The significance level was set at $p<0.05$. 

In our previous work \cite{bkheet2023using}, we explored temporal data obtained through action segmentation tools and examined its correlation with skill. In the current work, we extend our investigation into spatial data. Specifically, we introduce a novel metric to quantify the average distance a surgeon’s hands move during each unique gesture. This approach will facilitate the provision of more focused practice recommendations, honing in on gestures that require further refinement.

\subsection{2D Different Angles}
In order to investigate the influence of RGB camera angles on measurement accuracy, our approach involved analyzing the movement of hands from different angles, emphasizing how each angle uniquely captures aspects of the movement in 3D space. This approach underscores our aim to demonstrate the superiority of depth cameras, which provide 3D imagery, over RGB cameras that offer only 2D perspectives.
To accomplish this, we determined the geometric center of each hand for every frame. We then computed the [x, y, z] coordinates representing the hand's spatial position within the simulator's point cloud, a methodology previously established in \ref{methods_3d_path}.
These 3D coordinates were then projected onto three orthogonal 2D planes: XY, YZ, and XZ. 
This projection onto 2D planes serves to mimic the limited perspective of RGB cameras. By comparing these projections, we aim to highlight the constraints of 2D imaging in capturing the full complexity of hand movements in 3D space.

Subsequent to the projection, we quantified the distances covered by the hand within these 2D planes as if they were captured by a 2D camera. This comparison is critical for demonstrating that depth cameras, with their 3D imaging capabilities, provide a more comprehensive and accurate representation of hand movements in 3D space than 2D RGB cameras.


\section{Results}\label{Results}
\subsection{Object Detection}
This section presents the results of the YOLOv8 algorithm applied to object detection. Tables \ref{table:simulators_alltool} and \ref{table:simulators_onlyhands} provide a detailed overview of the algorithm's performance, specifically in terms of Average Precision (AP) for each class. The evaluation was conducted on two distinct models: one trained on RGB video data and the other on depth video data. These models were rigorously tested on a separate test set comprising 313 frames for the first simulator and 354 frames for the second from different participants, ensuring that the model's performance was evaluated on previously unseen data.

In the first simulator, the suture pad simulator, for the model trained on RGB video data using all the tools and the hands, we obtained $mAP_{50-95}(RGB)$ of 0.890. Similarly, for the same model trained on depth video data, the corresponding $mAP_{50-95}(Depth)$ was found to be 0.888.
For the models trained only on the hands, we obtained $mAP_{50-95}(RGB)$ of 0.976 and $mAP_{50-95}(Depth)$ of 0.963. These results highlight the consistency in the models' performance despite being trained on distinct data types, emphasizing the similar value that the depth camera gives us.

The second simulator, the fascia closure simulator, the model trained on RGB video data, exhibited performance with a $mAP_{50-95}(RGB)$ of 0.830. The model trained on depth video data for all the tools and the hands achieved a $mAP_{50-95}(Depth)$ of 0.801. 
For the models trained only on the hands, we obtained $mAP_{50-95}(RGB)$ of 0.945 and $mAP_{50-95}(Depth)$ of 0.966.

\begin{table}[h]
\centering
\resizebox{\textwidth}{!}
{
    \begin{tabular}{@{}lcccccccc@{}}
    \toprule
    &
    \multicolumn{3}{c}{Suture Pad Simulator}  &&  
    \multicolumn{3}{c}{Fascia Closure Simulator} 
    \\
    \cmidrule(lr){2-4} \cmidrule(lr){6-8} 
    Class &
    Occurrence &  $AP_{50-95}^{RGB}$ &  $AP_{50-95}^{Depth}$
    &&  Occurrence &  $AP_{50-95}^{RGB}$ &  $AP_{50-95}^{Depth}$ \\ 
    \midrule

Left Hand & 316   & 0.967 & 0.964 && 352 & 0.935 & 0.937  \\
Right Hand & 306   & 0.953 & 0.942 && 332 & 0.944 & 0.973  \\
Needle Driver   & 295   & 0.931 & 0.922 && 313 & 0.915 & 0.882  \\
Tissue Forceps       & 299   & 0.648 & 0.634 && 246 & 0.349 & 0.290  \\
Dressing Forceps       & 273   & 0.792 & 0.819 && 297 & 0.646 & 0.506  \\
Scissors        & 298   & 0.932 & 0.927 && 287 & 0.816 & 0.814  \\
Simulator       & 309   & 0.999 & 0.999 && 353 & 0.989 & 0.981  \\
\midrule
Average & - & 0.890  & 0.888  && - & 0.830 & 0.801  \\
   \bottomrule
\end{tabular}
}
\caption{Suture pad and fascia closure simulators - All Tools and Hands}
\label{table:simulators_alltool}
\end{table}

\begin{table}[h]
\centering
\resizebox{\textwidth}{!}{
\begin{tabular}{@{}lcccccccc@{}}
\toprule
  &
  \multicolumn{3}{c}{Suture Pad Simulator}  &&  
  \multicolumn{3}{c}{Fascia Closure Simulator} 
    \\
  \cmidrule(lr){2-4} \cmidrule(lr){6-8} 
  Class &
  Occurrence &  $AP_{50-95}^{RGB}$ &  $AP_{50-95}^{Depth}$
  &&  Occurrence &  $AP_{50-95}^{RGB}$ &  $AP_{50-95}^{Depth}$ \\ 
  \midrule

Left Hand & 316 & 0.980 & 0.965  &&  352  &  0.955 & 0.964 \\
Right Hand  & 306 & 0.972 & 0.961  && 332   & 0.933  & 0.968 \\
\midrule
Average &  - & 0.976  & 0.963  && -   & 0.945  & 0.966 \\
   \bottomrule
\end{tabular}
}
\caption{Suture pad and fascia closure simulators - Only Hands}
\label{table:simulators_onlyhands}
\end{table}

\subsection{Action Segmentation}
In the case of the Suture Pad simulator, models trained using depth features outperformed others across all evaluation measures, with the sole exception being UVAST's marginally higher edit score in the simple suture task. 
Using depth features, UVAST attained an accuracy of $78.22\%$ for the Simple Suture and $70.97\%$ for the Running Suture. At the same time, MS-TCN++ achieved accuracies of $76.75\%$ and $66.98\%$ for the same tasks, outperforming their respective RGB-based versions.

In the case of the Fascia Closure simulator, models trained using RGB showcase higher results across evaluation metrics, MS-TCN++ achieving an accuracy of $75.24\%$, and UVAST achieving an accuracy of $71.69\%$.
Nonetheless, as indicated in Table \ref{table:action_segmentation}, these results remain comparable to those achieved using depth features.

\begin{table}[h]
\centering
\resizebox{\textwidth}{!}{%
\begin{tabular}{@{}clrlllrlllrll@{}}
\toprule
&&
  \multicolumn{3}{c}{\textbf{Simple Suture}}  &&
  \multicolumn{3}{c}{\textbf{Running Suture}} &&
  \multicolumn{3}{c}{\textbf{Fascia}} \\ 
  \cmidrule(lr){3-5} \cmidrule(lr){7-9} \cmidrule(l){11-13} &&
  \textbf{F1@\{10, 25, 50\}} &  \textbf{Edit} &  \textbf{Acc} &   &
  \textbf{F1@\{10, 25, 50\}} &  \textbf{Edit} &  \textbf{Acc} &   &
  \textbf{F1@\{10, 25, 50\}} &  \textbf{Edit} &  \textbf{Acc} \\ \midrule
\multirow{2}{*}{\textbf{RGB}} &
MS-TCN++ &
    80.43 77.38 65.29 & 77.78 & 72.40 && 66.74 61.89 44.96 &  62.56 & 62.09 && 77.53 74.68 62.66 & 72.21 & 75.24 \\
  &  UVAST + Viterbi &
   83.84 81.39 67.63 & 81.57 & 74.92 && 69.94 65.65 52.04 & 70.26 & 64.73 && 73.66 70.41 59.33 & 69.47 & 71.69 \\ \midrule
\multirow{2}{*}{\textbf{Depth}} 
&  MS-TCN++ &
   82.63 80.40 69.56 & 79.13 & 76.75 && 71.96 68.02 52.33 & 67.83 & 66.98 && 70.22 66.71 53.33 & 63.36 & 67.20 \\
   &  UVAST + Viterbi &
   84.19 82.55 71.47 & 79.74 & 78.22 && 75.21 72.26 56.69 & 71.80 & 70.97 && 66.20 63.67 51.58 & 62.90 & 65.24 \\ \bottomrule
\end{tabular}%
}
\caption{Suture pad and fascia closure simulators - action segmentation results}
\label{table:action_segmentation}
\end{table}

\subsection{3D Hand Path Length}
\begin{figure}[hbt!]
    \centering
    \begin{subfigure}[b]{0.32\textwidth}
        \includegraphics[width=\textwidth]{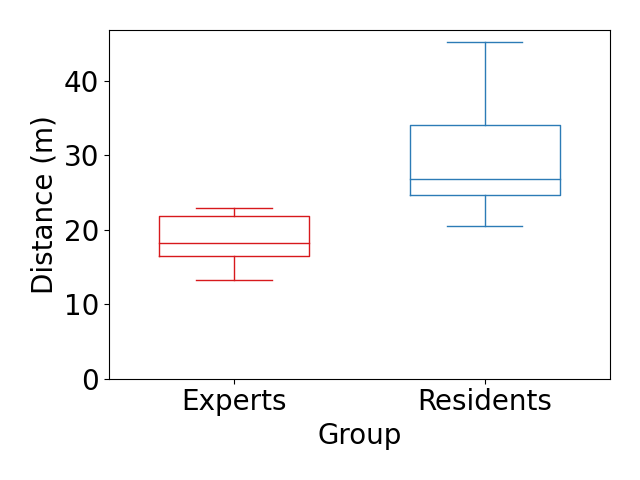}
        \caption{Simulator 1 - Task 1}
        \label{subfigure:hand_path_3d_simulator1_task1}
    \end{subfigure}%
    \hfill
    \begin{subfigure}[b]{0.32\textwidth}
        \includegraphics[width=\textwidth]{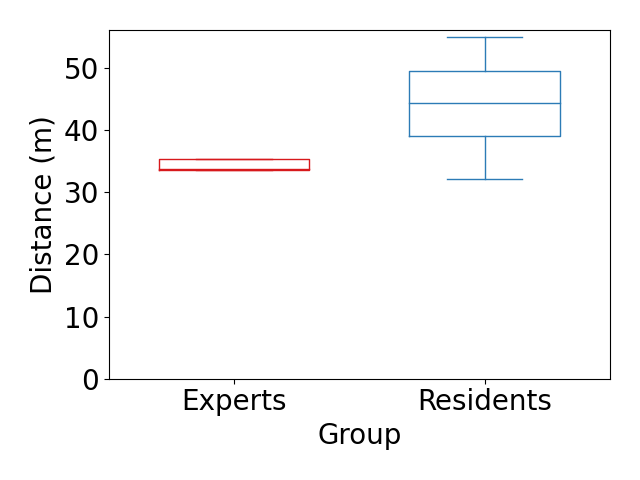}
        \caption{Simulator 1 - Task 2}
        \label{subfigure:hand_path_3d_simulator1_task2}
    \end{subfigure}%
    \hfill
    \begin{subfigure}[b]{0.32\textwidth}
        \includegraphics[width=\textwidth]{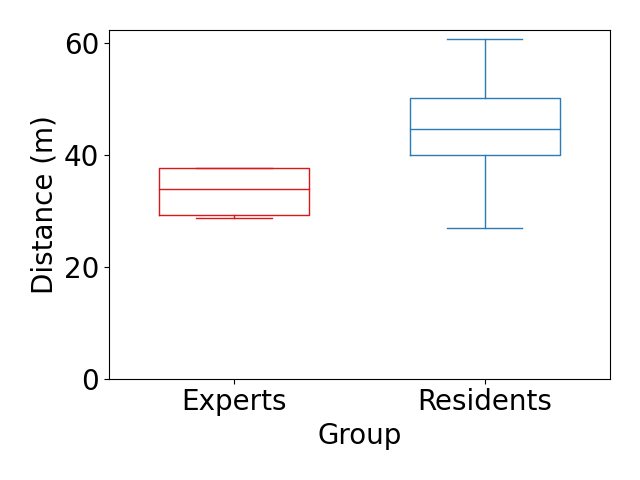}
        \caption{Simulator 1 - Task 3}
        \label{subfigure:hand_path_3d_simulator1_task3}        
    \end{subfigure}
    
    \begin{subfigure}[b]{0.32\textwidth}
        \includegraphics[width=\textwidth]{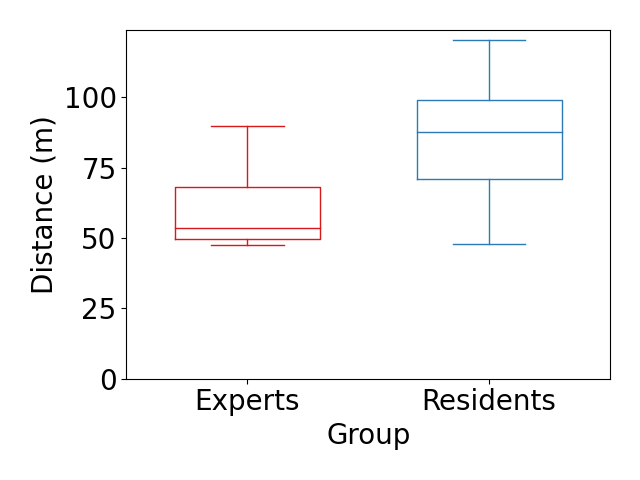}
        \caption{Simulator 1 - Task 4}
        \label{subfigure:hand_path_3d_simulator1_task4}
    \end{subfigure}
    \begin{subfigure}[b]{0.32\textwidth}
        \includegraphics[width=\textwidth]{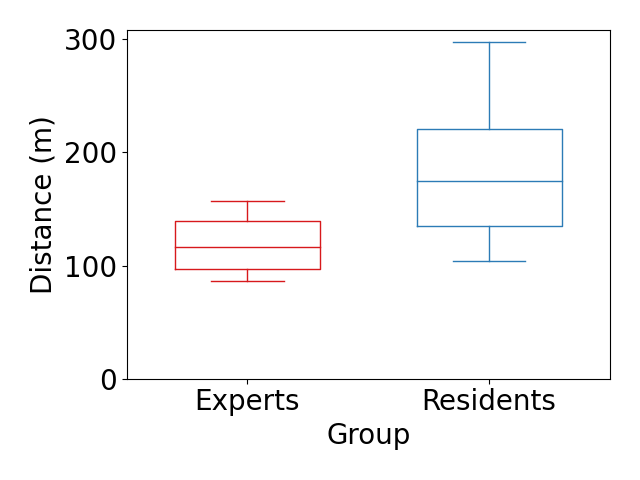}
        \caption{Simulator 2}
        \label{subfigure:hand_path_3d_simulator2}
    \end{subfigure} 
    \caption{Task-Based Box-plot of Hand Path Length for Residents and Experts}
\label{figure:hand_path_3d}
\end{figure}

As we expected, and shown for the 2D case in \cite{goldbraikh2022video}, the box plots in Fig. \ref{figure:hand_path_3d} reveal a consistent pattern across all tasks. Experts consistently navigated a shorter hand path compared to residents. 
In the suture pad simulator, for Task 1 - sub-figure \ref{subfigure:hand_path_3d_simulator1_task1}, the p-value of the Wilcoxon rank-sum test was 0.003, for Task 2 - sub-figure \ref{subfigure:hand_path_3d_simulator1_task2} it was 0.038, for Task 3 - sub-figure \ref{subfigure:hand_path_3d_simulator1_task3} the p-value was 0.021, for Task 4 - sub-figure \ref{subfigure:hand_path_3d_simulator1_task4} the p-value was 0.021, and for the fascia closure simulator two - sub-figure \ref{subfigure:hand_path_3d_simulator2} the p-value was 0.038. All the p-values are $p < 0.05$. 

These results indicate that the differences in hand path length between experts and residents are statistically significant, similar to \cite{lefor2020motion}, underscoring the value of expertise in surgical efficiency that can be captured using a depth camera.
Additionally, it's noteworthy that despite the small sample size in our datasets, we were able to achieve statistically significant p-values. This fact further reinforces the validity of our results, highlighting the robustness of our findings even with limited data.

\begin{figure}[h]
    \centering    
    \includegraphics[width=1\textwidth]{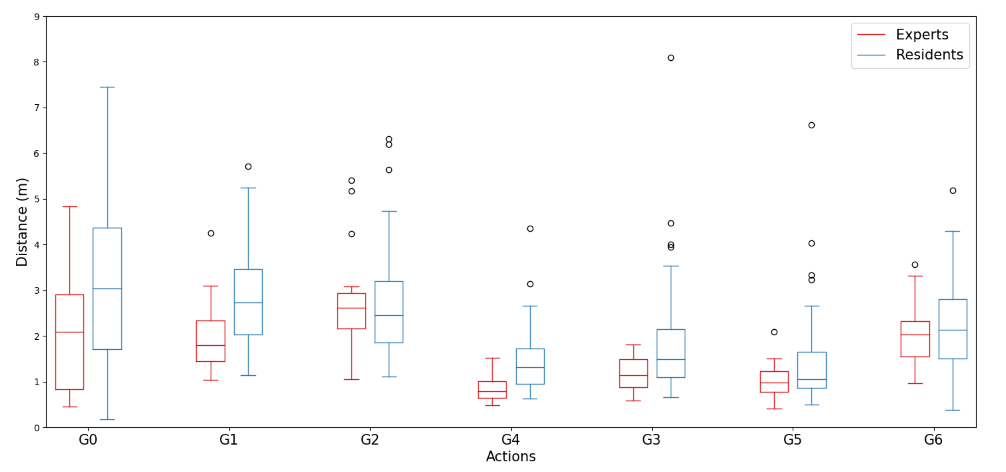}
    \caption{Gesture distance for experts and residents in the Simple Suture task}
    \label{fig:gesture_distance}
\end{figure}

Fig. \ref{fig:gesture_distance} serves as an initial exploration based on data collected from the Simple Suture simulator and gives us a more nuanced look at the hands' path length, showing which gestures require the most movement. Our results show a statistically significant difference in the distance passed when passing the needle ($p=0.001$), tying a knot ($p<0.001$), laying a knot ($p=0.008$), and holding the needle ($p=0.021$).
No significant difference was found when pulling the suture ($p=0.707$), cutting the suture ($p=0.056$), and for the distance moved when no action is performed ($p=0.283$).
This offers an initial validation for our proposed gesture distance metric.

\subsection{2D Different Angles}
Our data in Table \ref{figure:hand_path_2d_experts_residents} quantitatively confirmed these visual distortions, revealing a correlation between camera angle and measurement error. We found that certain angles disproportionately amplified or minimized specific types of movement, thereby providing an inaccurate representation of the true hand path. It becomes clear from these results that, because the hand's path distance in RGB video is always calculated along a 2D plane, results are inherently subject to variance due to camera angles. This issue can lead to a loss of up to a third of the actual, real-world data, as demonstrated in the XY case under Table \ref{figure:hand_path_2d_experts_residents}. This presents an inherent problem when using 2D cameras for applications that demand high accuracy and reliability.

In contrast, 3D cameras offer a solution to this issue by capturing the real-world position of the hands in a three-dimensional space, thereby eliminating the distortions introduced by varying camera angles. This allows for a more authentic and nuanced understanding of hand movements, as it captures the complete spatial relationships between different points in the hand’s path.

\begin{table}[h]
\centering
\resizebox{\textwidth}{!}{%
\begin{tabular}{@{}ccccccccccccccc@{}}
\toprule &&  \multicolumn{2}{c}{\textbf{XYZ}} &&  \multicolumn{2}{c}{\textbf{XY}} &&  \multicolumn{2}{c}{\textbf{YZ}} &&  \multicolumn{2}{c}{\textbf{XZ}} 
  \\ 
  \cmidrule(lr){3-4} \cmidrule(lr){6-7} \cmidrule(l){9-10} \cmidrule(l){12-13} &&
  \textbf{Experts} &  \textbf{Residents} &&  \textbf{Experts} &  \textbf{Residents} &&
  \textbf{Experts} &  \textbf{Residents} &&  \textbf{Experts} &  \textbf{Residents} \\ 
  \midrule
\multirow{4}{*}{\textbf{Suture Pad}}   
& Task 1 & 18.52 $\pm$ 3.51& 29.95 $\pm$ 7.59&& 11.00 $\pm$ 1.65& 17.77 $\pm$ 5.28&  & 16.06 $\pm$ 3.20 & 26.30 $\pm$ 6.88 && 17.43 $\pm$ 3.18 &  28.11 $\pm$ 7.16 \\
& Task 2 & 32.97 $\pm$ 6.76& 45.94 $\pm$ 11.86&& 19.36 $\pm$ 4.07& 27.69 $\pm$ 9.18 && 28.77 $\pm$ 6.03 & 40.17 $\pm$ 10.71&& 31.05 $\pm$ 6.41& 43.00 $\pm$ 11.04 \\ 
& Task 3 & 33.39 $\pm$ 3.88& 45.70 $\pm$ 11.98&& 19.38 $\pm$ 2.43 & 27.41 $\pm$ 8.04 && 29.41 $\pm$ 3.78& 40.28 $\pm$ 10.79 && 31.53 $\pm$ 3.53 & 42.45 $\pm$ 11.09\\ 
& Task 4 & 61.72 $\pm$ 15.71& 90.20 $\pm$ 26.10&& 57.83 $\pm$ 15.01 & 85.13 $\pm$ 23.85 && 19.38 $\pm$ 2.43 & 27.41 $\pm$ 8.04&& 29.41 $\pm$ 3.78 & 40.28 $\pm$ 10.79\\ 
\midrule
\multirow{1}{*}{\textbf{Fascia Closure}} 
&& 119.38 $\pm$ 27.64& 180.71 $\pm$ 52.59&& 74.39 $\pm$ 15.81& 125.18$\pm$ 37.06&& 102.37 $\pm$ 22.92& 154.63 $\pm$ 44.04 && 111.61$\pm$25.53 & 169.57$\pm$48.52 \\ \bottomrule
\end{tabular}
}
\caption{Hand Path Length in the Planes XY, YZ and XZ (Meter) divided to experts and residents\centering}
\label{figure:hand_path_2d_experts_residents}
\end{table}

\section{Discussion and Conclusion}\label{Discussion}
Depth cameras have several advantages over traditional RGB cameras.
These advantages include resilience to lighting and positional variations, simplified data compression, and enhanced privacy. In this study, we assessed if they provide a viable and competent alternative to RGB cameras for tasks like object detection and action segmentation in surgical environments.

Analysis of motion using depth data revealed statistically significant differences between experts and residents. Additionally, our study into the effects of camera angles on measurement accuracy indicates that depth cameras are less susceptible to variations in camera position and setup. Consequently, readily available and affordable depth cameras could offer a new, accessible approach for self-directed and independent training, coupled with objective feedback.

It is essential to acknowledge the limitations of our study. The data set size is crucial when using deep learning tools and statistical tests. More accurate results could have been achieved with the availability of a more extensive dataset.
Due to the high memory demands of the Viterbi algorithm, we had to use a simplified version. This is a balanced approach and a common solution in applications involving the Viterbi algorithm. While the full algorithm might provide slightly more precision in certain cases, the simplified version aligned well with our research requirements. Also, while a larger dataset could potentially offer finer details, rigorous methods were employed to ensure the validity of our study given the available data.

In conclusion, our research contributes to the field of surgical skill assessment. By championing the adoption of depth cameras, we provide a more accurate, privacy-conscious, and robust approach to evaluating surgical proficiency. The advantages of depth cameras, combined with our empirical findings, underscore their potential to alter how surgical skills are assessed and trained, offering a solid foundation for future advancements in this domain.

\backmatter

\bmhead{Acknowledgments}
This research was funded by a grant from the University of Wisconsin-Madison Department of Surgery.
The authors would like to thank Abbie Schafer for assisting in data annotation.

\section*{Declarations}
\bmhead{Conflict of interest} The authors declare that they have no conflict of
interest.

\bmhead{Ethical approval} This study was granted an exemption by the University of Wisconsin-Madison Institutional Review Board.

\bmhead{Informed consent} Informed consent was obtained from all individual
participants included in the study.
\bibliography{sn-bibliography}
\end{document}